\documentclass{article}
\usepackage{ijcai20}

\usepackage{times}

\usepackage{soul}
\usepackage{url}
\usepackage[hidelinks]{hyperref}
\usepackage[utf8]{inputenc}
\usepackage[small]{caption}
\usepackage{graphicx}
\usepackage{amsmath}
\usepackage{booktabs}
\usepackage{amsthm}
\usepackage{amssymb}
\usepackage{subfigure}
\usepackage{algorithm}
\usepackage{algorithmic}

\title{On Deep Unsupervised Active Learning}

\author{
Changsheng Li$^1$\and
Handong Ma$^2$\and
Zhao Kang$^2$\and
Ye Yuan$^1$\and
Xiao-Yu Zhang$^3$\And
Guoren Wang$^1$\footnote{Guoren Wang is the corresponding author.}\\
\affiliations
$^1$School of Computer Science and Technology, Beijing Institute of Technology, Beijing, China\\
$^2$SCSE, University of Electronic Science and Technology of China, Chengdu, China\\
$^3$Institute of Information Engineering, Chinese Academy of Sciences, Beijing, China\\
\emails
\{lcs, yuan-ye\}@bit.edu.cn,
201921080133@std.uestc.edu.cn,
zkang@uestc.edu.cn,
zhangxiaoyu@iie.ac.cn,
wanggrbit@126.com
}

\begin{document}

\maketitle

\begin{abstract}
Unsupervised active learning has attracted increasing attention in recent years, where its goal is to select representative samples in an unsupervised setting for human annotating. Most existing works are based on shallow linear models by assuming that each sample can be well approximated by the span (i.e., the set of all linear combinations) of certain selected samples, and then take these selected samples as representative ones to label. However, in practice, the data do not necessarily conform to linear models, and how to model nonlinearity of data often becomes the key point to success. In this paper, we present a novel Deep neural network framework for Unsupervised Active Learning, called DUAL. DUAL can explicitly learn a nonlinear embedding to map each input into a latent space through an encoder-decoder architecture, and introduce a selection block to select representative samples in the the learnt latent space. In the selection block, DUAL considers to simultaneously preserve the whole input patterns as well as the cluster structure of data. Extensive experiments are performed on six publicly available datasets, and experimental results clearly demonstrate the efficacy of our method, compared with state-of-the-arts.
\end{abstract}

\section{Introduction}
In many real-world applications, there are lots of available unlabeled data whereas labeled data are often difficult to get. It is expensive and time-consuming to manually annotate the data, especially when domain experts must be involved. In this situation, active learning provides a promising way to reduce the cost by automatically selecting the most informative or representative samples from an unlabeled data pool for human labeling. In other words, these selected samples can improve the performance of the model (e.g., classifier) the most if they are labeled and used as training data. Due to its huge potential, active learning has been successfully applied to various tasks such as image classification \cite{joshi2009multi}, object detection \cite{vijayanarasimhan2014large}, video recommendation \cite{cai2019multi}, etc.

Currently, research works on active learning follow two lines according to whether supervised information is involved \cite{Li2017Joint}.
The first line concentrates on how to leverage data structures to select representative samples in an unsupervised manner. Typical algorithms include  transductive experimental design (TED) \cite{Kai2006Active},  locally linear reconstruction  \cite{Lijun2011Active}, robust representation and structured sparsity \cite{Nie2013Early}, joint active learning and feature selection (ALFS) \cite{Li2017Joint}.
The other line considers the problems of querying informative samples.
Such methods basically need a pre-trained classifier to select samples, which means that they need some initially labeled data for training.
In this line,
many approaches have been proposed in the past decades \cite{Freund1997Selective,kapoor2007selective,Jain2009Active,Huang2010Active,Elhamifar2013A,Zheng2015Querying,Zhang2017Bidirectional,haussmann2019deep}.  Due to
the limitation of space, we refer the reader to \cite{aggarwal2014active} and \cite{settles2009active} for more details. In this paper, we focus on unsupervised active learning, since it is a challenging problem because of the lack of supervised information.

Most existing works on unsupervised active learning \cite{Kai2006Active,Nie2013Early,hu2013active,fy201510,shi2016diversifying,li2017active,Li2017Joint} assume that each data point can be reconstructed by the span, i.e., the set of all linear combinations, of a selected sample subset, and resort to shallow linear models to minimize the reconstruction error.
Such methods often suffer from the following limitations:
first, they attempt to reconstruct the whole dataset, but ignore the cluster structure of data.
For example, let us consider an extreme case: assume that there are 100 samples, among which 99 samples are positive and one sample is negative. Thus, the negative sample is very important, and it should be selected as one of the representative samples. However, if we only minimize the total reconstruction loss of all samples by a selected sample subset, then it is very likely that the negative sample is not selected, because of it being far from the positive samples.
The second limitation of these methods is that they use shallow and linear mapping functions to reveal the intrinsic structure of data, thus may fail in handling data with  complex (often nonlinear) structures.
To address this issue, the manifold adaptive experimental design (MAED) algorithm \cite{cai2011manifold} attempts to learn the representation in a manifold adaptive kernel space obtained by incorporating the manifold structure into the reproducing kernel Hilbert space (RKHS).
As we know, kernel based methods  heavily depend on the choice of kernel functions, while it is difficult to find an optimal or even suitable kernel function in real-world applications.

To overcome the above limitations, in this paper, we propose a novel unsupervised active learning framework based on a deep learning model, called Deep Unsupervised Active Learning (DUAL). DUAL takes advantage of an encoder-decoder model to explicitly learn a nonlinear latent space, where DUAL can perform sample selection by introducing a \emph{selection} block at the junction between the encoder and the decoder. In the selection block, we attempt to simultaneously reconstruct the whole dataset and the cluster centroids of the data. Since the selection block is differentiable, DUAL is an end-to-end trainable framework. To the best of our knowledge, our approach constitutes the first attempt to select representative samples based on a deep neural network in an unsupervised setting. Compared with existing unsupervised active learning approaches, our method significantly differs from them in the following aspects:
\begin{itemize}
    \item DUAL directly learns a nonlinear representation by a multi-layer neural network, which offers stronger ability to discover nonlinear structures of data.
    \item In contrast to kernel-based approaches, DUAL can provide explicit transformations, avoiding subjectively choosing the kernel function in advance.
    \item
    Not only can DUAL model the whole input patterns by the selected representative samples, but also it can preserve cluster structures of data well.
\end{itemize}

We extensively evaluate our method on six publicly available datasets. The experimental results show our DUAL outperforms the related state-of-the-art methods.
\section{Related Work}
As mentioned above, we focus on the studies on unsupervised active learning. In this section, we briefly review some algorithms devoted to unsupervised active learning and some related works to this topic.

\subsection{Unsupervised Active Learning}
Among existing unsupervised active learning methods, the most representative one is the transductive experimental design (TED) \cite{Kai2006Active}, where its goal is to select samples that can best represent the dataset using a linear representation. Thus, TED proposes an objective function as:
\begin{align}\label{ted}
 & \min_{\mathbf{Z,S}} \sum_{i=1}^n(||\mathbf{x}_i - \mathbf{Z}\mathbf{s}_i ||_2^2 + \alpha ||\mathbf{s}_i||_2^2) \nonumber \\
 s.t. \ \   \mathbf{S}= & [\mathbf{s}_1,...,\mathbf{s}_n] \in \mathbb{R}^{m\times n}, \mathbf{Z}= [\mathbf{z}_1,...,\mathbf{z}_m]\subset \mathbf{X},
\end{align}
where  $\mathbf{X}=[\mathbf{x}_1,...,\mathbf{x}_n] \in \mathbb{R}^{d\times n}$ denotes the data matrix, where $d$ is the dimension of samples and $n$ is the number of samples.
$\mathbf{Z}\subset \mathbf{X}$ denotes the selected sample subset, and $\mathbf{S}$ is the reconstruction coefficient matrix. $\alpha$ is a tradeoff parameter to control the amount of shrinkage. $||\cdot||_2$ denotes the $l_2$-norm of a vector.

Following TED, more unsupervised active learning methods have been proposed in recent years.
Inspired by the idea of Locally Linear Embedding (LLE) \cite{roweis2000nonlinear},
\cite{Lijun2011Active} propose to represent each sample by a linear combination of its neighbors, with the purpose of preserving the intrinsic local structure of data.
Similarly, ALNR also incorporates the neighborhood relation into the sample selection process, where the nearest neighbors are expected to have stronger effect on the reconstruction of  samples \cite{hu2013active}.
\cite{Nie2013Early} extend TED to a convex formulation by introducing a structured sparsity-inducing norm, and take advantage of a robust sparse representation loss function to suppress outliers.
\cite{shi2016diversifying} extend convex TED to a diversity version for selecting  complementary samples.
More recently, ALFS \cite{Li2017Joint} study the coupling effect on unsupervised active learning and feature selection, and perform them jointly via the CUR matrix decomposition \cite{mahoney2009cur}.
The above methods are linear models, which cannot model nonlinear structures of data in many read-world scenarios.
Thus, \cite{cai2011manifold} propose a kernel-based method to perform nonlinear sample selection in a manifold adaptive kernel space.
However, how to choose appropriate kernel functions for kernel-based methods is usually unclear in practice.

Unlike these  approaches, our method explicitly learns a nonlinear embedding by a deep neural network architecture, so that the nonlinear structures of data can be discovered in the latent space, and thus a better representative sample subset can be obtained.

\subsection{Matrix Column Subset Selection}
Unsupervised active learning is related to one popular mathematical problem: matrix column subset selection (MCSS) \cite{chan1987rank}. The MCSS problem aims to select a subset of columns from an input matrix, such that the selected columns can capture as much of the input as possible.
More precisely, it attempts to select $m$ columns of $\mathbf{X}$ to form a new matrix $\mathbf{Z}\in \mathbf{R}^{d\times m}$ that minimizes the following residual:
\begin{align}
     \min_{\mathbf{Z,C}} ||\mathbf{X} - \mathbf{Z}\mathbf{C} ||_{\varepsilon} =  ||\mathbf{X} - \mathbf{Z}\mathbf{Z}^{\dagger}\mathbf{X} ||_{\varepsilon} \nonumber
\end{align}
where $\mathbf{Z}^{\dagger}$ is the Moore-Penrose pseudoinverse of $\mathbf{Z}$. $\mathbf{Z}\mathbf{Z}^{\dagger}$ denotes the projection onto the $m$-dimensional space spanned by the columns of $\mathbf{Z}$.
$\varepsilon = 2$ or $F$ denotes the spectral norm or Frobenius norm.


Different from our method, MCSS still attempts to reconstruct the input based on a linear combination of the selected columns, which cannot model the nonlinearity of data.

\begin{figure*}
\centering
\includegraphics[width=0.8\linewidth]{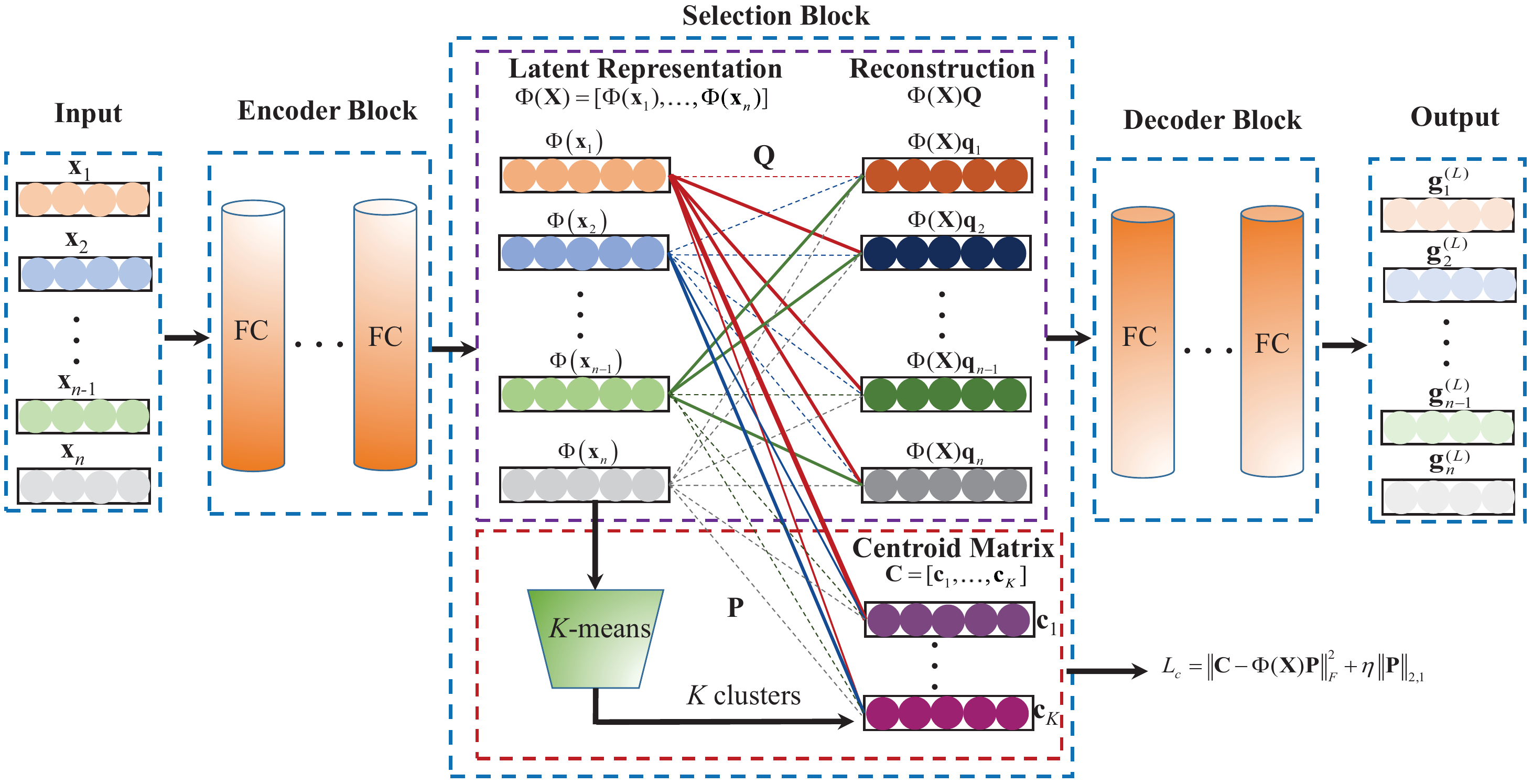}
\caption{
Illustration of the overall architecture. DUAL consists of an encoder block, a selection block and a decoder block.
In particular, the encoder and decoder blocks are used to learn a nonlinear representation. The selection block consists of two branches, where each is made
up of one fully connected layer without bias and nonlinear activation functions.
The top branch attempts to capture as much of the latent representation $\Phi(\mathbf{X})$ as possible, while the bottom one aims to approximate the $K$ cluster centroids well. The dash line in the fully connected
layer of the selection block denotes that the sample in the left side has no contribution to the reconstruction of the corresponding sample in the right side, i.e., its weight in $\mathbf{Q}$ or $\mathbf{P}$ is equal to zero.
}
\label{framework}
\end{figure*}

\section{Deep Unsupervised Active Learning}
In this section, we will elaborate the details of the proposed DUAL model for unsupervised active learning.
As shown in Figure \ref{framework}, DUAL mainly consists of the following three blocks: an encoder block and a decoder block are used to learn a nonlinear representation, and a selection block at the junction between the encoder and the decoder used for selecting samples. Before explaining how DUAL is specifically designed for these parts, we first give our problem setting.

\subsection{Problem Setting}
Let $\mathbf{X} = [\mathbf{x}_1 , ..., \mathbf{x}_n ] \in \mathbb{R}^{d\times n}$ denote a data matrix, where $d$ and $n$ are the feature dimension and the number of training data, respectively.  Our goal is to learn a nonlinear transformation $\Phi$ to map input $\mathbf{X}$ to a new latent representation $\Phi(\mathbf{X})$, and then select $m$ samples expected to  not only approximate the latent representation $\Phi({\mathbf{X}})$ well but also preserve the cluster structure of the training data.
This problem is quite challenging, since solving it exactly is a hard combinatorial optimization problem (NP-hard).
After obtaining a sample subset based on our method DUAL, a prediction model (e.g., a SVM classifier for classification ) will be trained by labeling the selected $m$ samples and using $\Phi(\cdot)$ as their new feature representations.

\subsection{Encoder Block}
In order to learn the nonlinear mapping $\Phi$, we utilize a deep  neural network to map each input to a latent space. Since we have no access to the labeled data, we adopt an encoder-decoder architecture because of its effectiveness for unsupervised learning. Our encoder block consists of $L+1$ layers for performing $L$ nonlinear transformations as the desired $\Phi$ (The detail of the decoder block is introduced in Section \ref{decoder}).
For easy of presentation, we first provide the definition of the output of each layer in the encoder block:
\begin{align}\label{hidden}
\mathbf{h}^{(l)}_i = \sigma (\mathbf{W}_e^{(l)} \mathbf{h}^{(l-1)}_i + \mathbf{b}_e^{(l)}), l=1,\ldots,L,
\end{align}
where $\mathbf{h}^{(0)}_i=\mathbf{x}_i, i=1,\cdots,n$, denotes the original training data $\mathbf{X}$ as the input of the encoder block.
$\mathbf{W}_e^{(l)}$ and $\mathbf{b}_e^{(l)}$ are the weights and bias associated with the $l$-th hidden layer, respectively.
$\sigma(\cdot)$ is a nonlinear activation function.
Then, we can define our latent representation $\Phi(\mathbf{X})$ as:
\begin{align}
\Phi(\mathbf{X}) =\mathbf{H}^{(L)}= [\mathbf{h}^{({L})}_1,\cdots,\mathbf{h}^{({L})}_n] \in \mathbb{R}^{d'\times n}
\end{align}
where $d'$ denotes the dimension of our latent representation.

\subsection {Selection Block}
As discussed above,  we aim to seek a sample subset which can better capture the whole input patterns and simultaneously preserve the cluster structure of data.
To this end, we introduce a selection block at the junction between the encoder and the decoder, as shown in Figure \ref{framework}.
In Figure \ref{framework}, the selection block consists of two branches, of which each is composed of one fully connected layer but without bias and nonlinear activation functions. The top branch is used to select a sample subset to approximate all samples in the latent space. The bottom one aims to reconstruct the cluster centroids, such that the cluster structure can be well preserved.
Next, we will introduce the two branches in detail.

\textbf{Top branch:} In order to best approximate all samples,  we present to minimize the following loss function:
\begin{align}\label{nphard}
    \mathcal{L}_a= \sum_{i=1}^n (||\Phi(\mathbf{x}_i) - \Phi(\mathbf{Y})\mathbf{q}_i||_{\ell}^2 + \gamma ||\mathbf{q}_i||_1) \\
    s.t. \ \ \Phi(\mathbf{Y}) = [\Phi(\mathbf{y}_1), \ldots, \Phi(\mathbf{y}_m)]\in \mathbb{R}^{d'\times m} \nonumber
\end{align}
where $\Phi(\mathbf{Y}) \subset \Phi(\mathbf{X})$ denotes the selected $m$ samples, and $\mathbf{q}_i \in \mathbb{R}^m$ is the reconstruction coefficients for sample $\Phi(\mathbf{x}_i)$.
$||\cdot||_1$ denotes the $l_1$-norm of a vector.
$||\cdot||_{\ell}$ denotes the $\ell$-norm of a vector, indicating certain loss measuring strategy. In this paper, we use a common norm, the $l_2$-norm, for simplicity. $\gamma \geq 0$ is a  tradeoff parameter.

The first term in Eq. (\ref{nphard}) aims to pick out $m$ samples to reconstruct the whole dataset  in the latent space, while the second term is a regularization term to enforce the coefficient sparse.
Unfortunately, there is not an easy solution to (\ref{nphard}), as it is a  combinatorial optimization. Inspired by \cite{Li2017Joint}, we relax (\ref{nphard}) to  a convex optimization problem, and write it into a matrix format as
\begin{align}\label{top}
 \min_{\mathbf{Q}\in \mathbb{R}^{n \times n}} \ \mathcal{L}_a &=  ||\Phi(\mathbf{X}) - \Phi(\mathbf{X}) \mathbf{Q}||_F^2 + \gamma ||\mathbf{Q}||_{2,1} \nonumber\\
 s.t.&\ ||diag(\mathbf{Q})||_1=0
\end{align}
where $||\cdot||_{2,1}$ denotes the $l_{2,1}$-norm of a matrix, defined as sum of the $l_2$-norms of row vectors.
The constraint condition ensures the diagonal elements of $\mathbf{Q}$ equal to zeros, avoiding the degenerated solution.
To minimize (\ref{top}), we utilize the fact that,  $\mathbf{Q}=[\mathbf{q}_1,\ldots,\mathbf{q}_n]$ can be thought of the parameters of a fully connected layer without bias and nonlinear activations, such that $\Phi$ and $\mathbf{Q}$ can be solved jointly through a standard backpropagation procedure.

\textbf{Bottom branch:} In the top branch, the selected samples can approximate the whole dataset well, while it might fail to preserve the cluster structure of data. To solve this problem, we first utilize a clustering algorithm to cluster the data into some clusters in the latent space, and then select a sample subset to best approximate the obtained cluster centroids.

Specifically, given the latent representation $\Phi(\mathbf{X}) $, we can obtain $K$ clusters based on a clustering algorithm, $K$-means used in this paper. Then, we denote the  cluster centroid matrix $\mathbf{C}$ as
\begin{align}
\mathbf{C} = [\mathbf{c}_1,\mathbf{c}_2,\ldots, \mathbf{c}_K] \in \mathbb{R}^{d' \times K}
\end{align}
where $\mathbf{c}_k$ denotes the $k$-th cluster centroid, and $K$ is the number of clusters.

In order to preserve the cluster structure, we minimize the following loss function:
\begin{align}\label{cluloss}
\mathcal{L}_c = ||\mathbf{C} - \Phi(\mathbf{X})\mathbf{P}||^2_F + \eta ||\mathbf{P}||_{2,1}
\end{align}
where $\mathbf{P}$ is the coefficient matrix for reconstructing $\mathbf{C}$.
$\eta \geq 0$ is a  tradeoff parameter.
Similarly, $\mathbf{P}$ can also be thought of the parameters of a fully connected layer without bias and nonlinear activation functions, and thus can be
optimized jointly with $\Phi$.

\subsection{Decoder Block}\label{decoder}

As aforementioned, since we have no access to the label information,  we attempt to recover each input by a decoder block, to guide the learning of the latent representation. In other words, each input actually plays the role of a supervisor.
Similar to the encoder block, our decoder block also consists of $L+1$ layers for performing $L$ nonlinear transformations. The output of each layer in the decoder block can be defined as:
\begin{align}\label{hidden2}
\mathbf{g}^{(l)}_i = \sigma (\mathbf{W}_d^{(l)} \mathbf{g}^{(l-1)}_i + \mathbf{b}_d^{(l)}), l=1,\ldots,L,
\end{align}
where $\mathbf{g}^{(0)}_i=\Phi(\mathbf{X})\mathbf{q}_i$ means that the output of the selection block is used as the input of the decoder block.
$\mathbf{W}_d^{(l)}$ and $\mathbf{b}_d^{(l)}$ are the weights and bias associated with the $l$-th hidden layer in the decoder block, respectively.

Then, the reconstruction loss function is defined as:
\begin{align}\label{recloss}
    \mathcal{L}_r= \sum_{i=1}^n||\mathbf{x}_i - \mathbf{g}_i^{(L)}||^2_2
    = ||\mathbf{X} - \mathbf{G}^{(L)}||_F^2,
\end{align}
where  $\mathbf{g}_i^{(L)}$ denotes the output of the decoder to the input $\mathbf{x}_i$, and $\mathbf{G}^{(L)}$ is expressed by
$\mathbf{G}^{(L)} = [\mathbf{g}_1^{(L)}, \mathbf{g}_2^{(L)},\ldots, \mathbf{g}_n^{(L)}]$.

\subsection{Overall Model and Training}
After introducing all the building blocks of this work, we now give the final training objective and explain how to jointly optimize it.
Based on the Eq. (\ref{top}), (\ref{cluloss}), and (\ref{recloss}),  the final loss function is defined as:
\begin{align}\label{total loss}
\min \ \mathcal{L}= \mathcal{L}_r + \alpha \mathcal{L}_a + \beta \mathcal{L}_c
\end{align}
where $\alpha$ and $\beta$ are two positive tradeoff parameters.

In (\ref{total loss}), there are three reconstruction loss terms. The first term denotes the reconstruction loss of the encoder-decoder model in Eq. (\ref{recloss}). The second term corresponds to input patterns reconstruction loss in Eq. (\ref{top}). The last term is the cluster centroids reconstruction loss as shown in Eq. (\ref{cluloss}).

To solve  (\ref{total loss}), we present a three-stage training strategy which is an end-to-end trainable fashion. Firstly, we pre-train the encoder and decoder block in the beginning without considering the selection block.
After that,  we utilize the output of the encoder block as the latent representation to perform $K$-means, and regard the obtained $K$ cluster centroids as the centroid matrix $\mathbf{C}$ for subsequent sample selection.
Lastly, we use the pre-trained parameters to initialize the encoder and decoder blocks, and load all data into a batch to optimize the whole network, i.e., minimizing the loss (\ref{total loss}).
Throughout the experiment, we use three fully connected layers in the encoder and decoder blocks, respectively. The rectified linear unit (ReLU) is used as the non-linear activation function.
In addition, we use Adam \cite{kingma2014adam} as the optimizer, where the learning rate is set to $1.0\times10^{-4}$.
\begin{figure*}
\centering
\subfigure[Urban]{\includegraphics[width=0.3\linewidth]{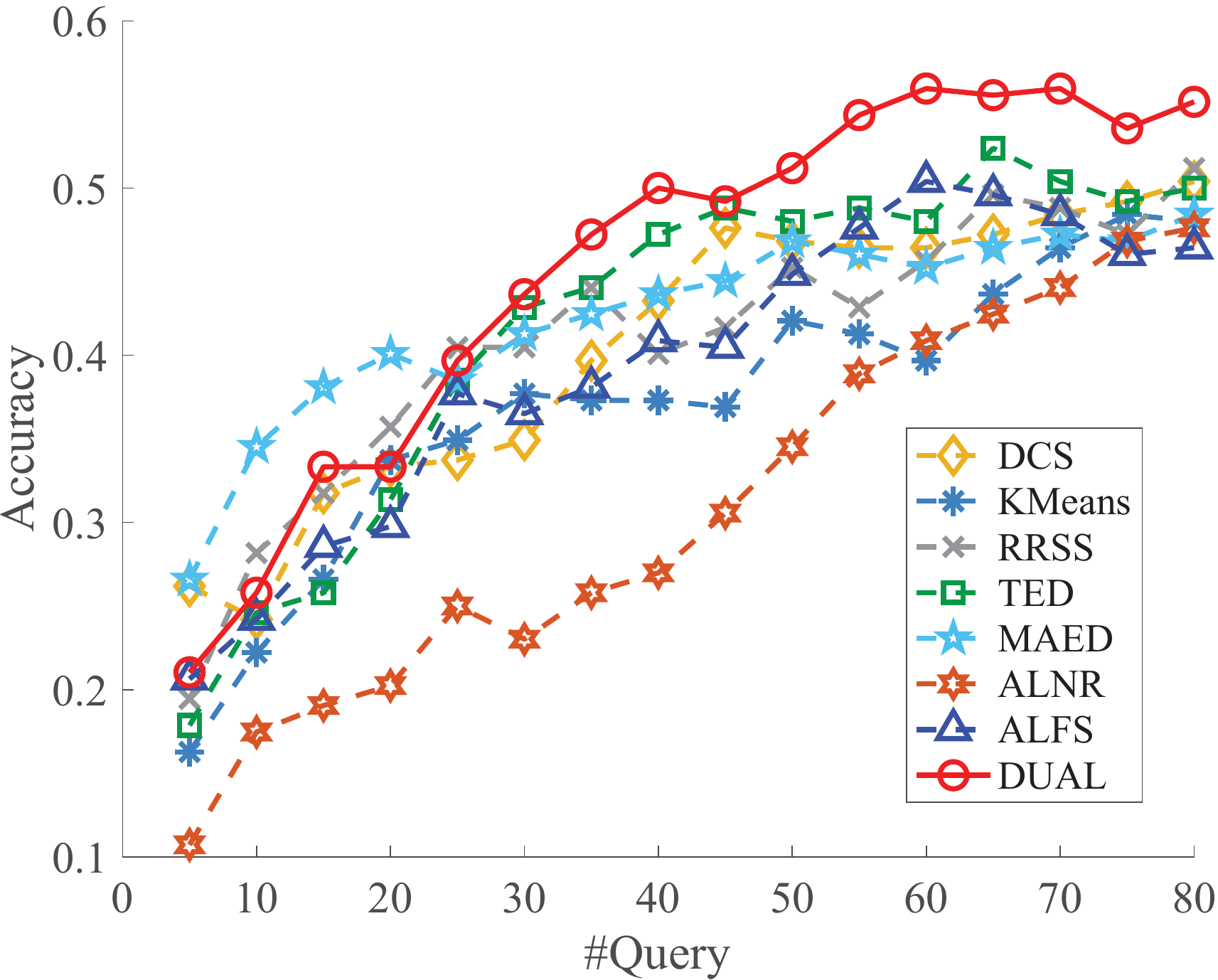}}
\subfigure[MASS]{\includegraphics[width=0.3\linewidth]{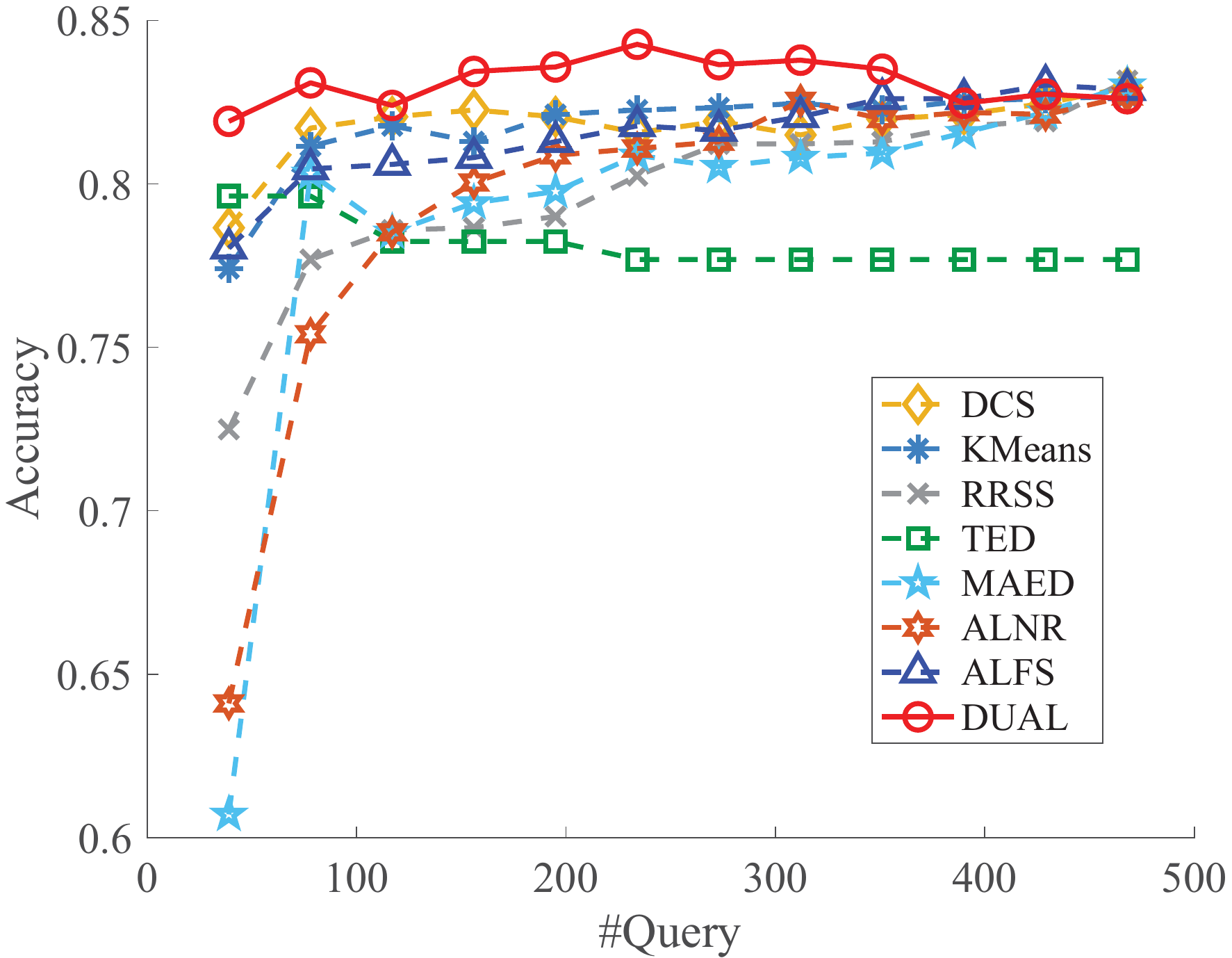}}
\subfigure[Plant Species Leaves]{\includegraphics[width=0.3\linewidth]{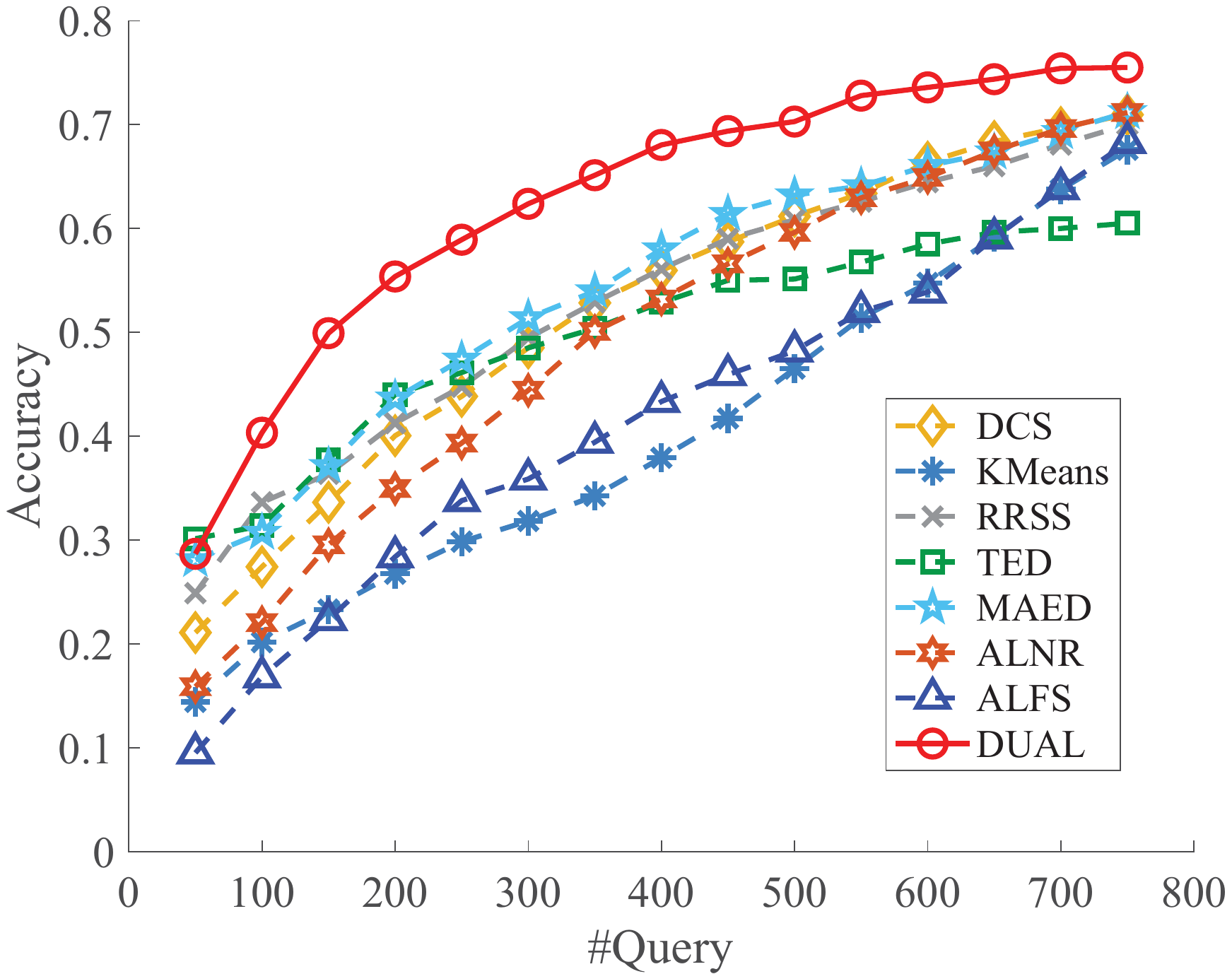}}
\subfigure[sEMG]{\includegraphics[width=0.3\linewidth]{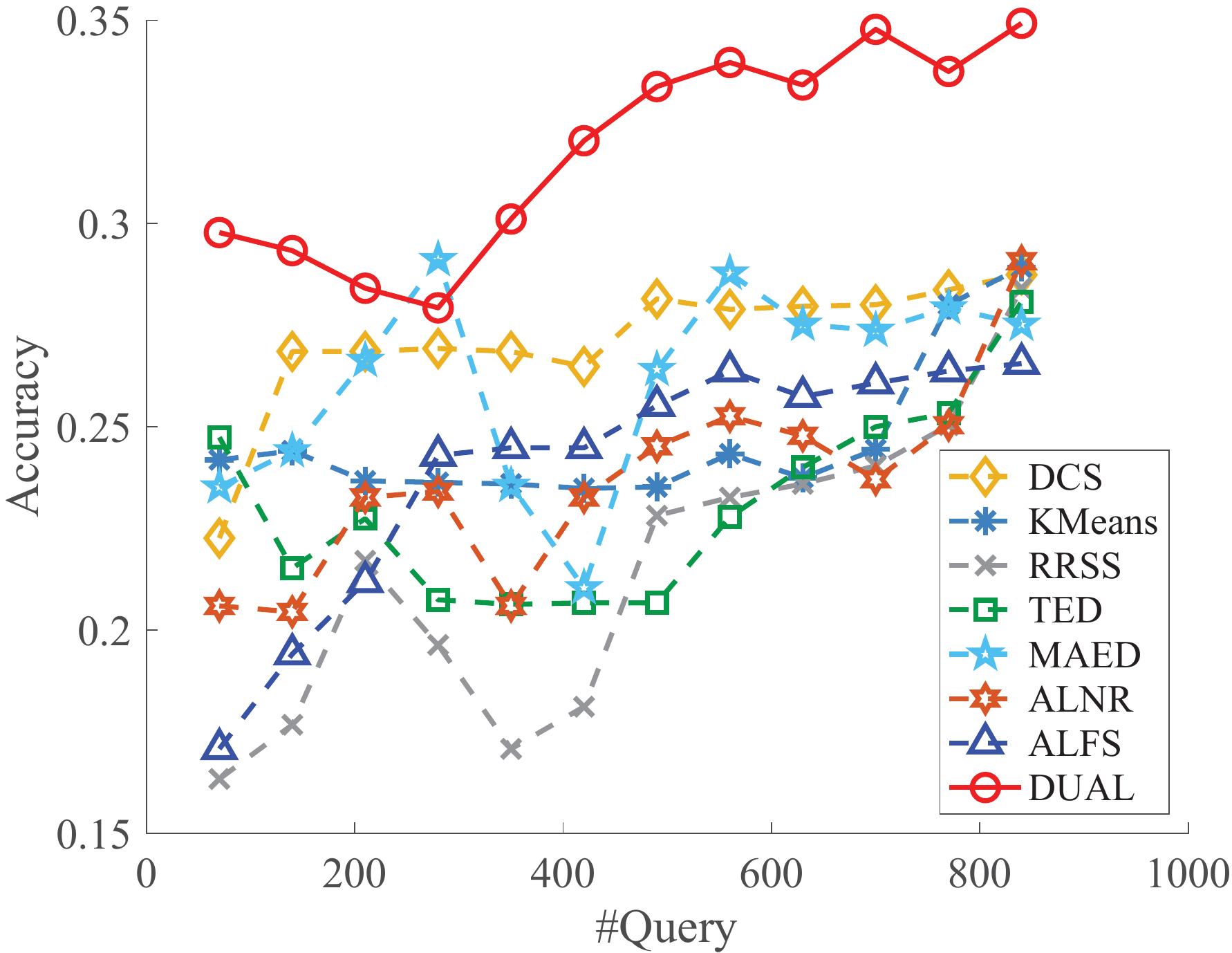}}
\subfigure[Waveform]{\includegraphics[width=0.3\linewidth]{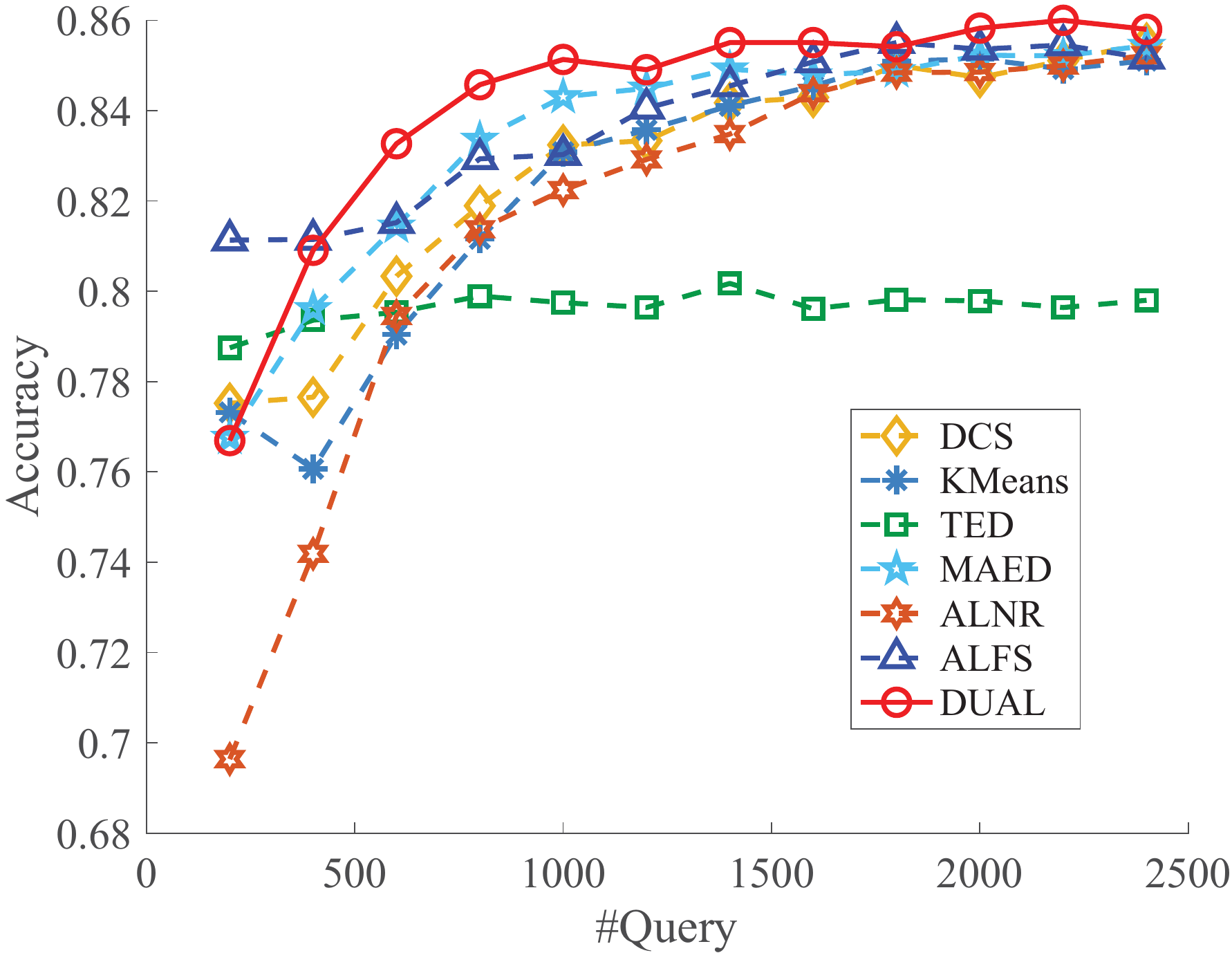}}
\subfigure[Gesture Phase Segment]{\includegraphics[width=0.3\linewidth]{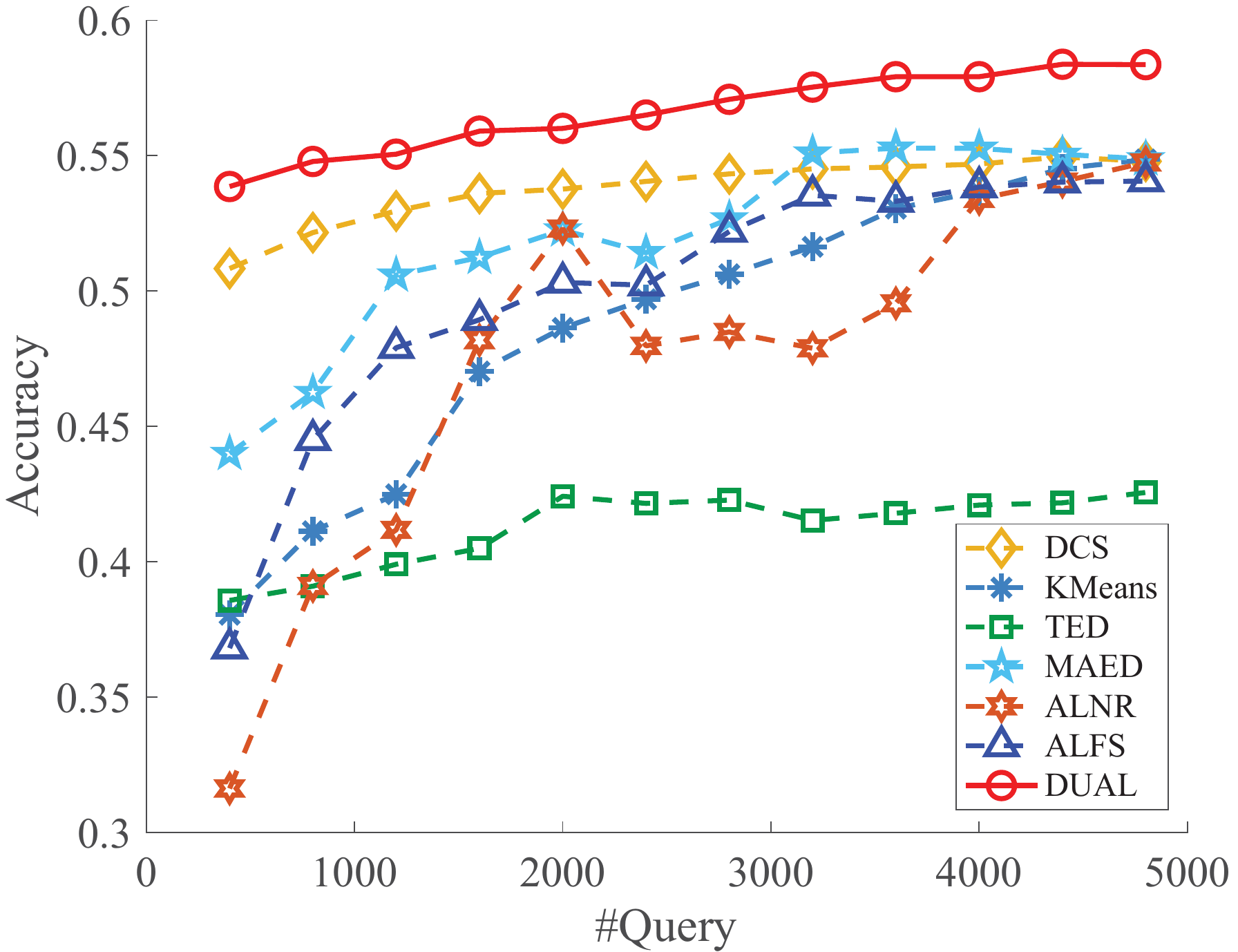}}
\caption{ Comparisons of different active learning methods in terms of accuracy on six benchmark datasets.}
\label{accuracy}
\end{figure*}

Once the model is trained, we can obtain two reconstruction coefficient matrices $\mathbf{Q}$ and $\mathbf{P}$. We then use a simple strategy to select the most representative samples based on $\mathbf{Q}$ and $\mathbf{P}$. Specifically, we calculate the $l_2$-norm of the rows of $\mathbf{Q}$ and $\mathbf{P}$ respectively, and obtain two corresponding vectors $\hat{\mathbf{q}}$ and $\hat{\mathbf{p}}$. After that, we normalize
$\hat{\mathbf{q}}$ and $\hat{\mathbf{p}}$ to the range of [0,1].
Finally, we can sort all the samples by adding normalized $\hat{\mathbf{q}}$ to normalized $\hat{\mathbf{p}}$ in descending order, and select the top $m$ samples as the most representative ones.

\section{Experiments}
 To verify the effectiveness of our method DUAL, we perform the experiments on six publicly available datasets which are widely used for active learning \cite{baram2004online}. The details of these datasets are summarized in Table \ref{dataset} \footnote{ These datasets are downloaded from the UCI Machine Learning Repository}.
 \begin{table}
\centering
\begin{tabular}{|c|c|c|c|}
\hline
 {Dataset}   & Size     & Dimension   & Class   \\
\hline
Urban Land Cover  &   168 &    148   &    9   \\\hline
Waveform &    5000 &   40   &   3  \\ \hline
sEMG  &   1800 &    3000   &    6   \\\hline
Plant Species Leaves &    1600 &   64   &   100  \\ \hline
Gesture Phase Segment  &   9873 &    50   &    5   \\\hline
Mammographic Mass  &    961 &   6   &   2  \\ \hline
\end{tabular}
\caption{Summary of Experimental Datasets.}
\label{dataset}
\end{table}

\subsection{Experimental Setting}

\textbf{Compared methods}\footnote{All source codes are obtained from the  authors of the corresponding papers, except $K$-means and ALNR.}: We compare DUAL with several typical unsupervised active learning algorithms, including TED \cite{Kai2006Active}, RRSS \cite{Nie2013Early}, ALNR \cite{hu2013active}, MAED \cite{cai2011manifold}, ALFS \cite{Li2017Joint}.
We also compare Deterministic Column Sampling(DCS) \cite{papailiopoulos2014provable} in our experiments.
 In addition, we also take $K$-means as a baseline.

\textbf{Experimental protocol}: Following \cite{Li2017Joint}, for each dataset, we randomly select 50$\%$ of the samples as candidates for sample selection, and use the rest as the testing data.
To evaluate the effectiveness of sample selection, we  train a SVM classifier with a linear kernel and  $C=100$ by using these selected samples as the training data.  We set the parameters $\gamma=\eta$ for simplicity, and search all tradeoff parameters in our algorithm from $\{0.01, 0.1, 1, 10\}$. The number of clusters $K$ is searched from $\{5, 10, 20, 50\}$.
We use accuracy and AUC to measure the performance.
We repeat every test case five times, and report the average result.

\subsection{Experimental Result}
\begin{figure*}
\centering
\subfigure[Urban]{\includegraphics[width=0.3\linewidth]{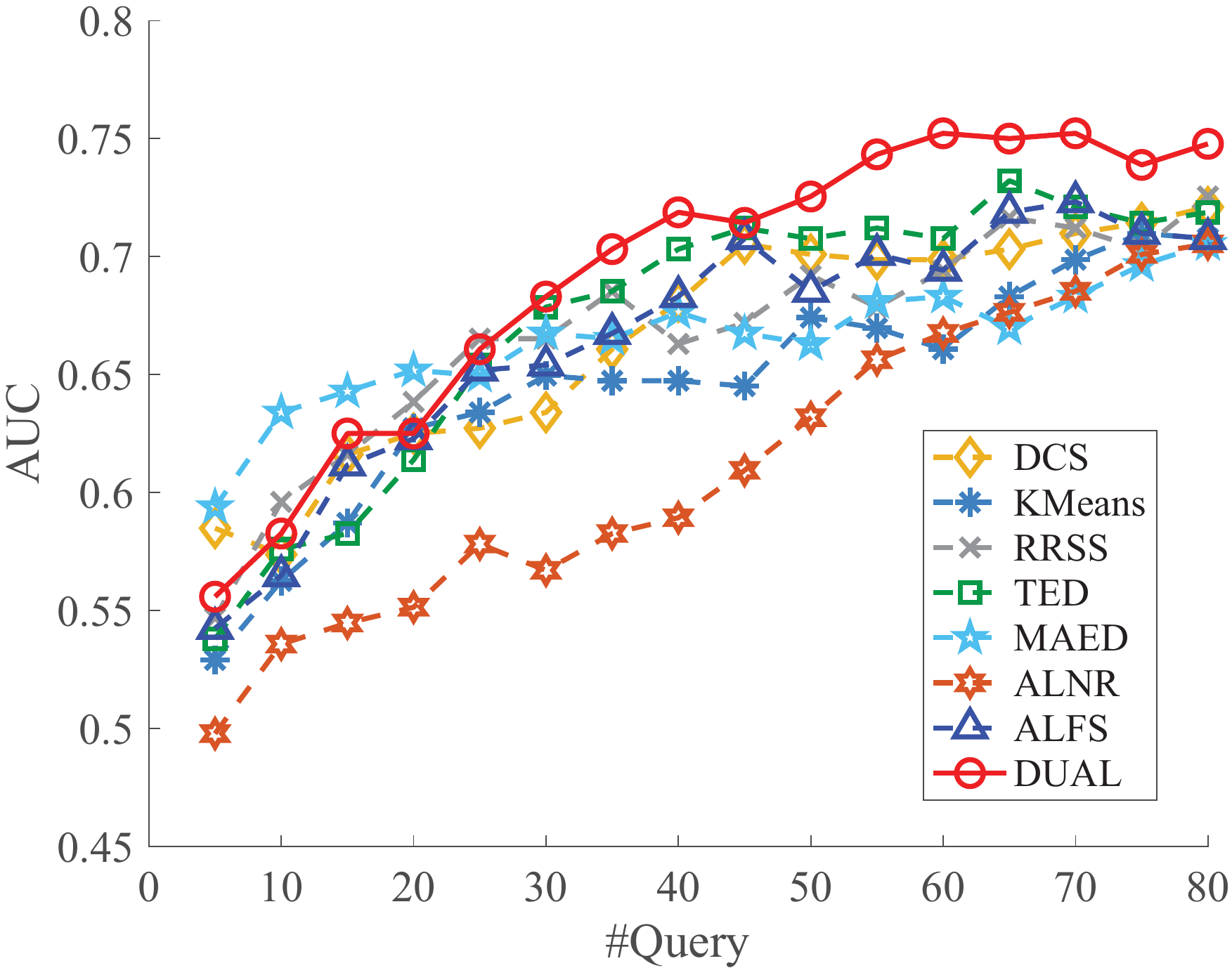}}
\subfigure[MASS]{\includegraphics[width=0.3\linewidth]{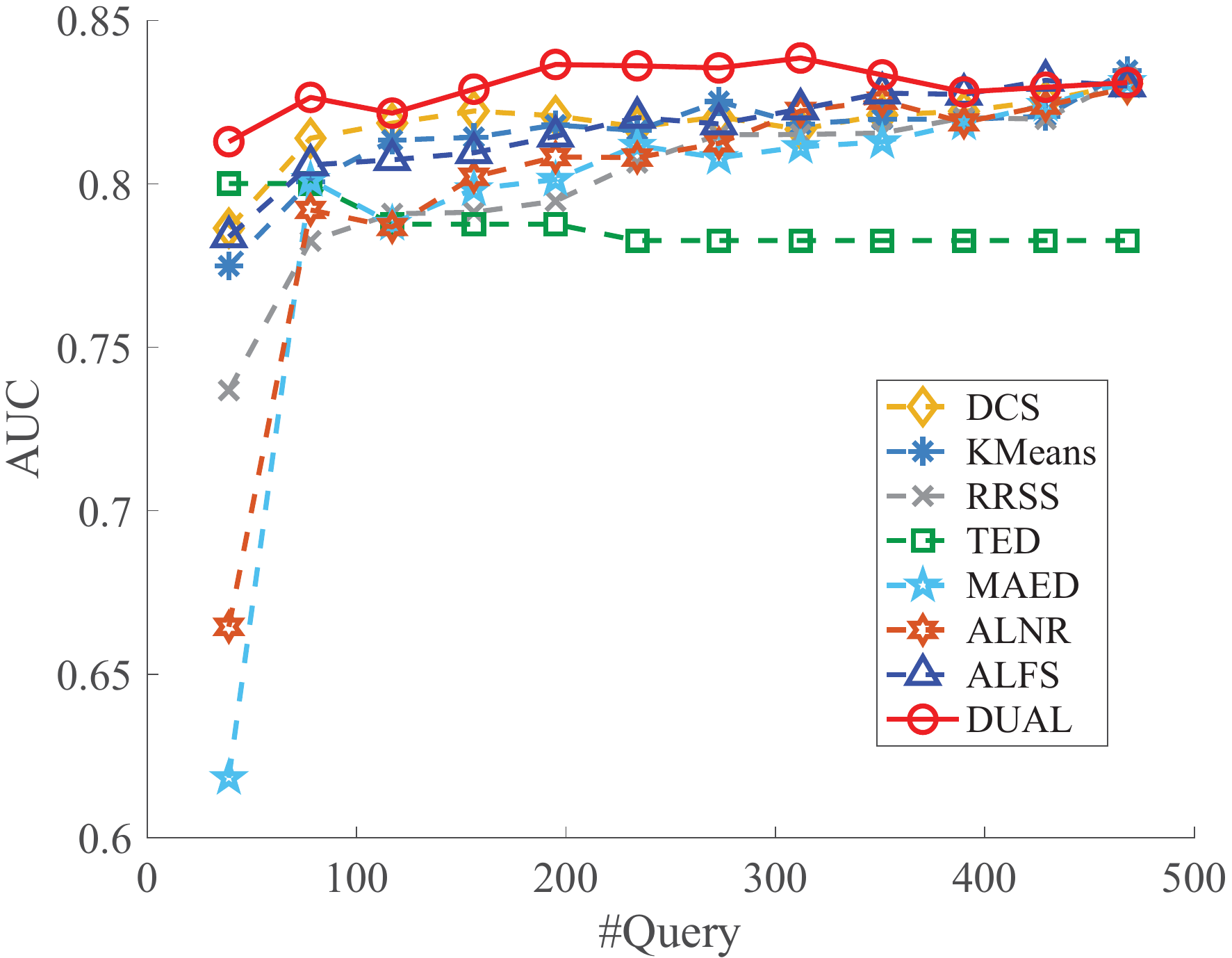}}
\subfigure[Plant Species Leaves]{\includegraphics[width=0.3\linewidth]{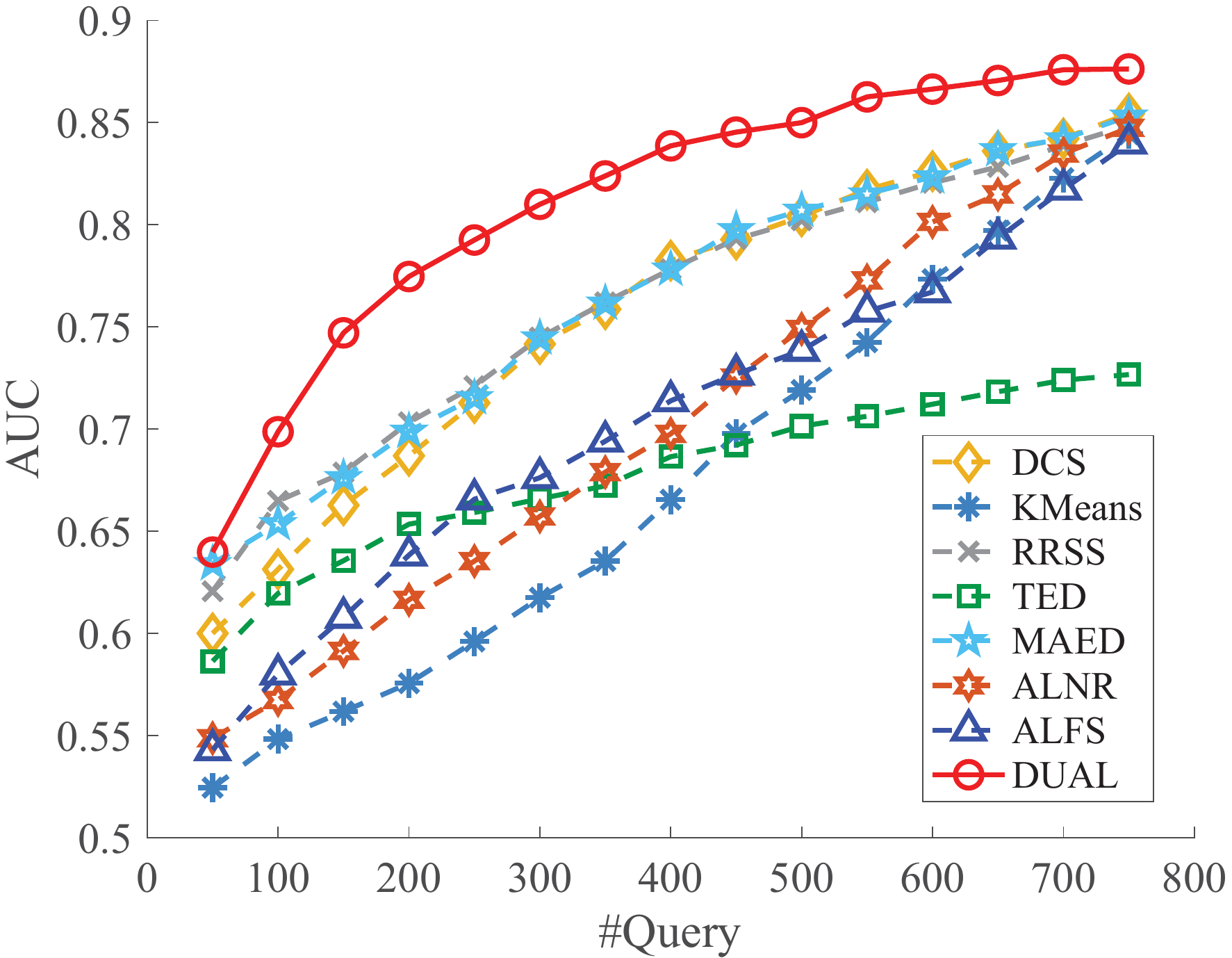}}
\subfigure[sEMG]{\includegraphics[width=0.3\linewidth]{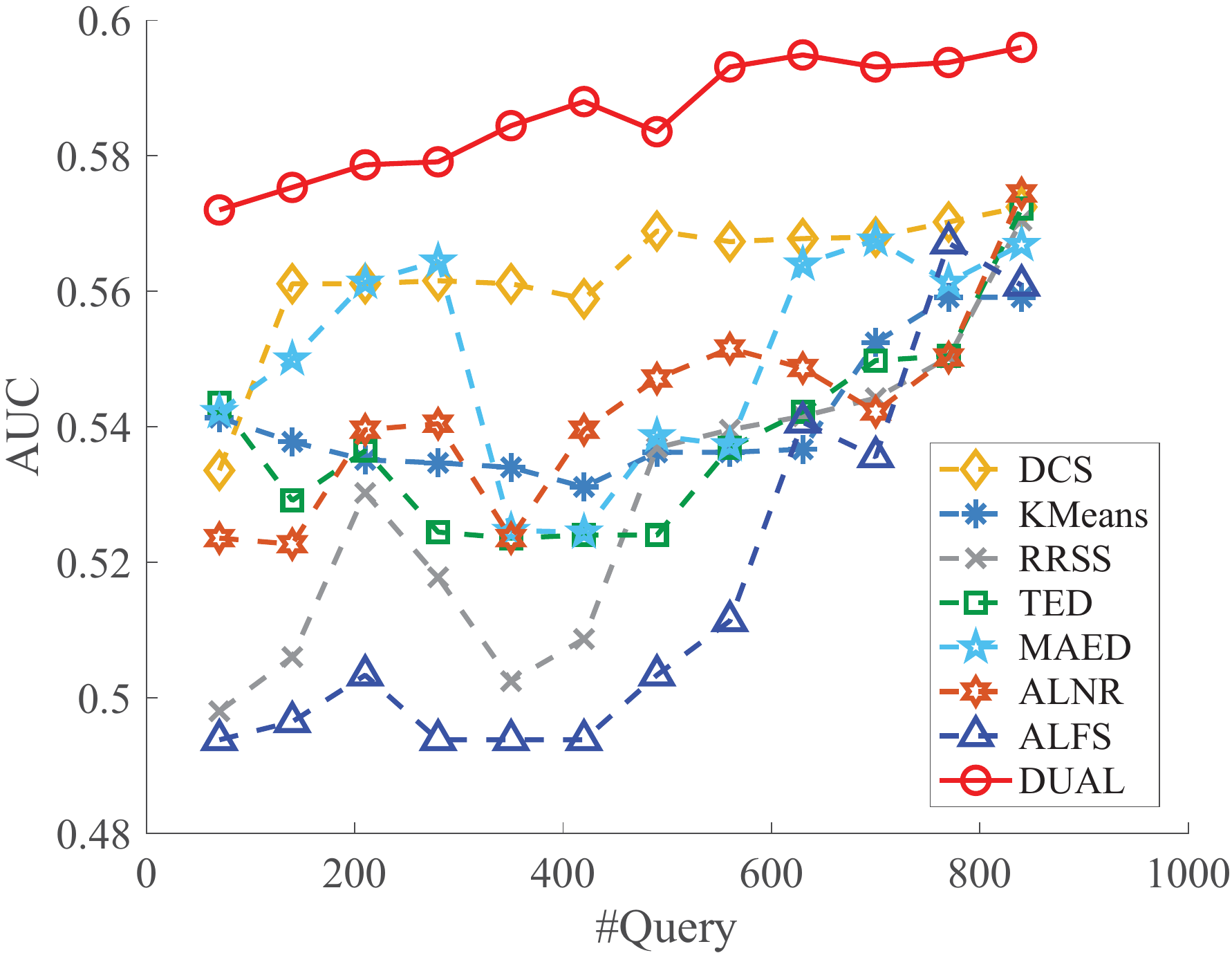}}
\subfigure[Waveform]{\includegraphics[width=0.3\linewidth]{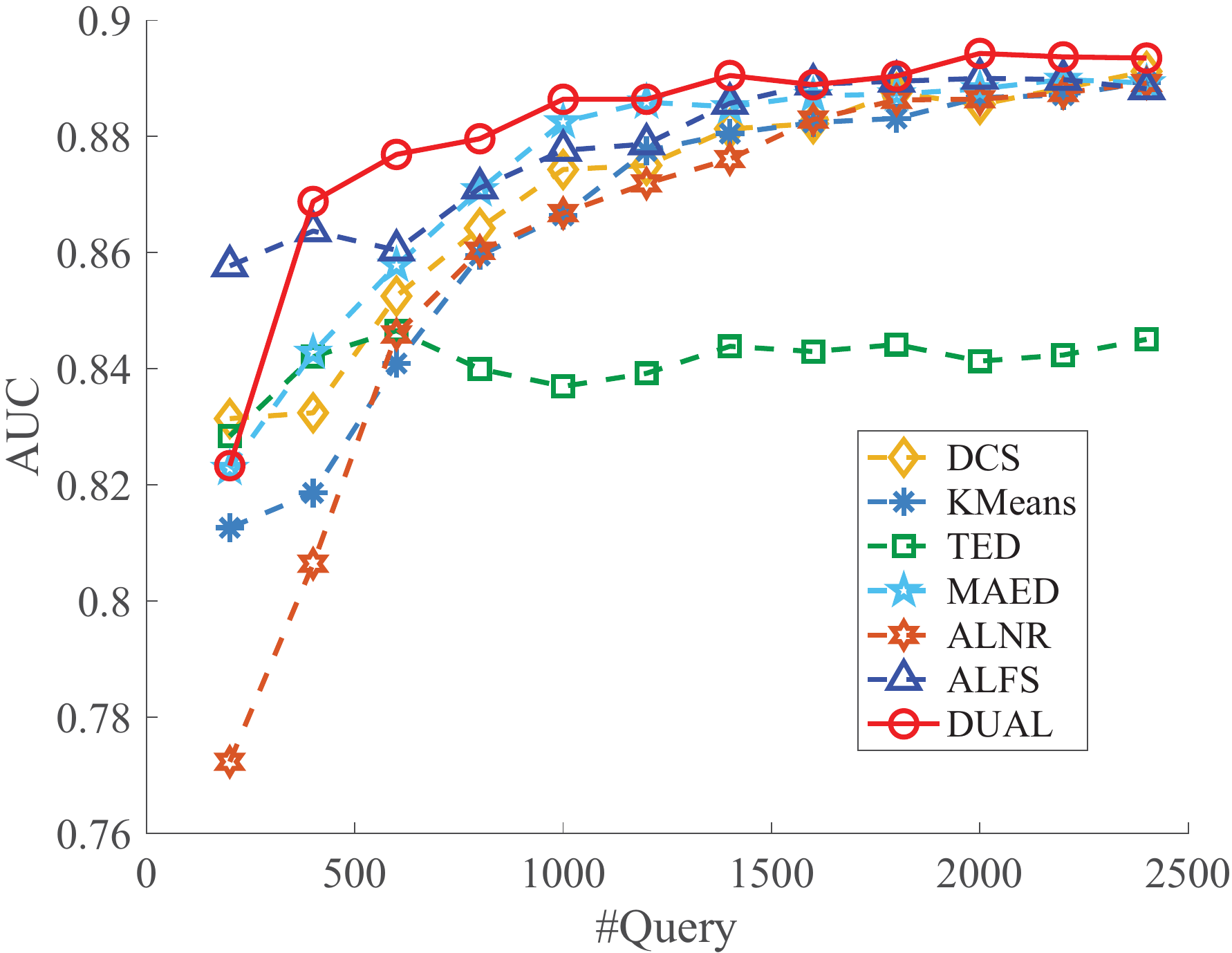}}
\subfigure[Gesture Phase Segment]{\includegraphics[width=0.3\linewidth]{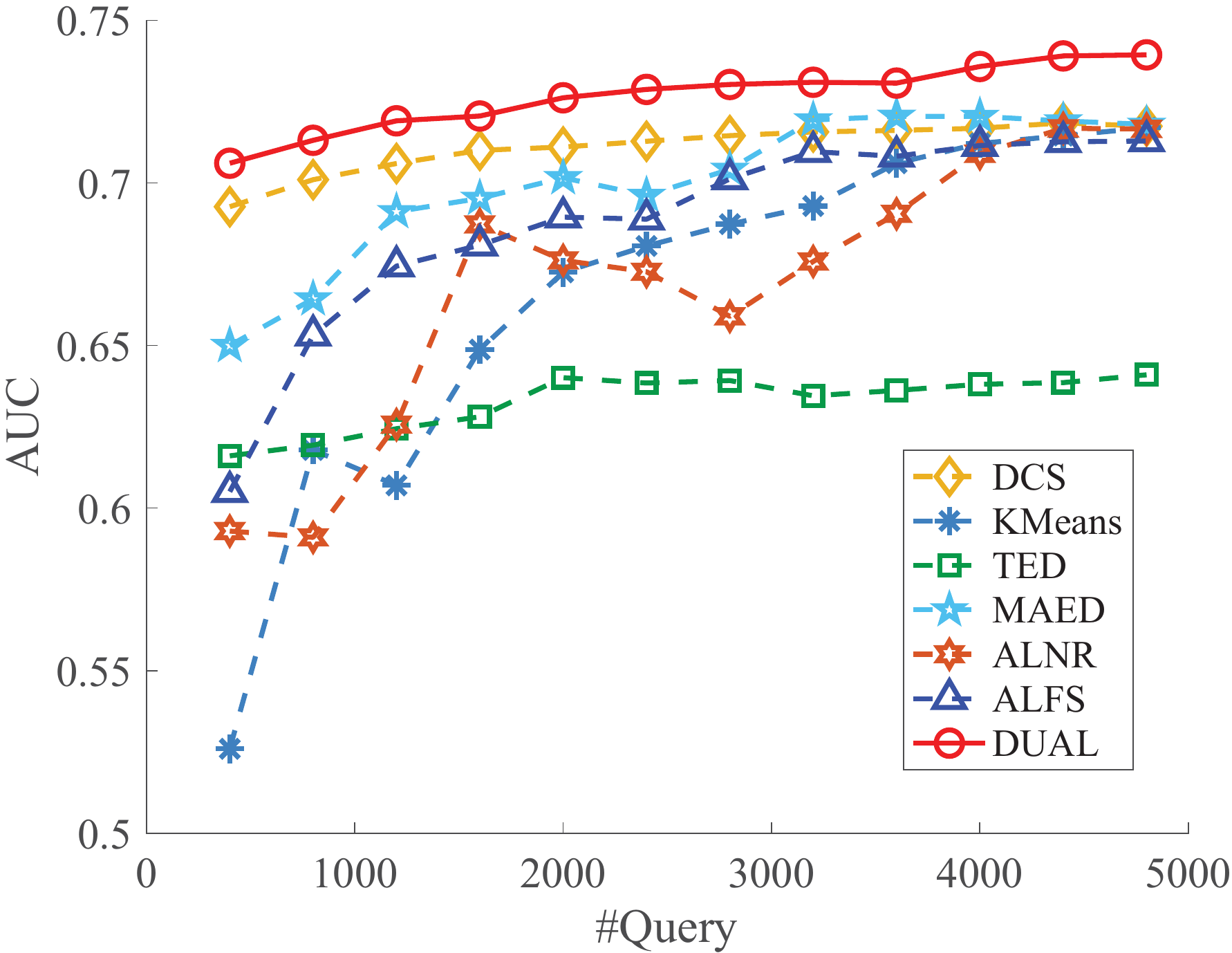}}
\caption{ Comparisons of different active learning methods in terms of AUC on six benchmark datasets.}
\label{auc}
\end{figure*}
\begin{figure}
\centering
\subfigure[Urban]{\includegraphics[width=0.42\linewidth]{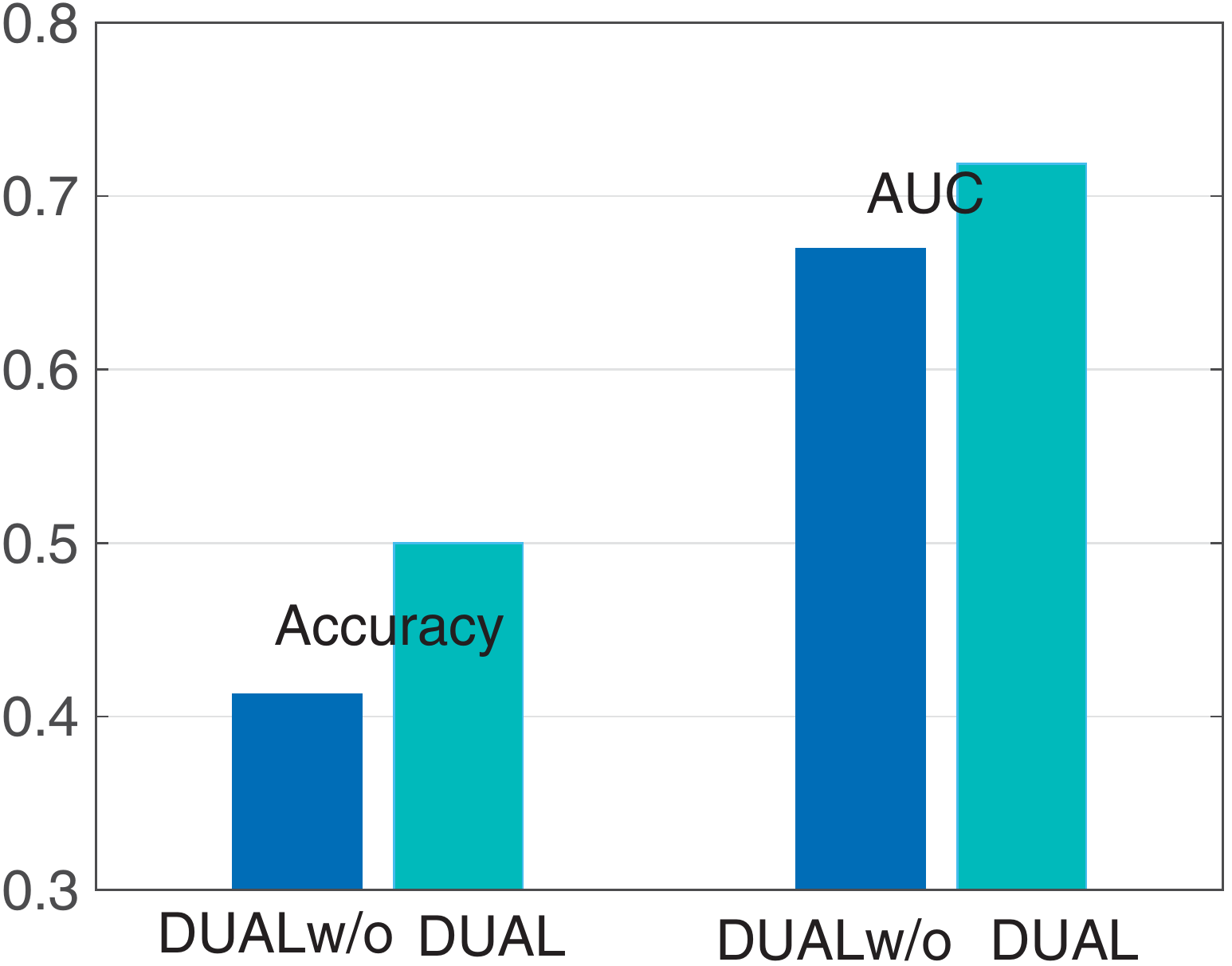}}
\subfigure[Waveform]{\includegraphics[width=0.42\linewidth]{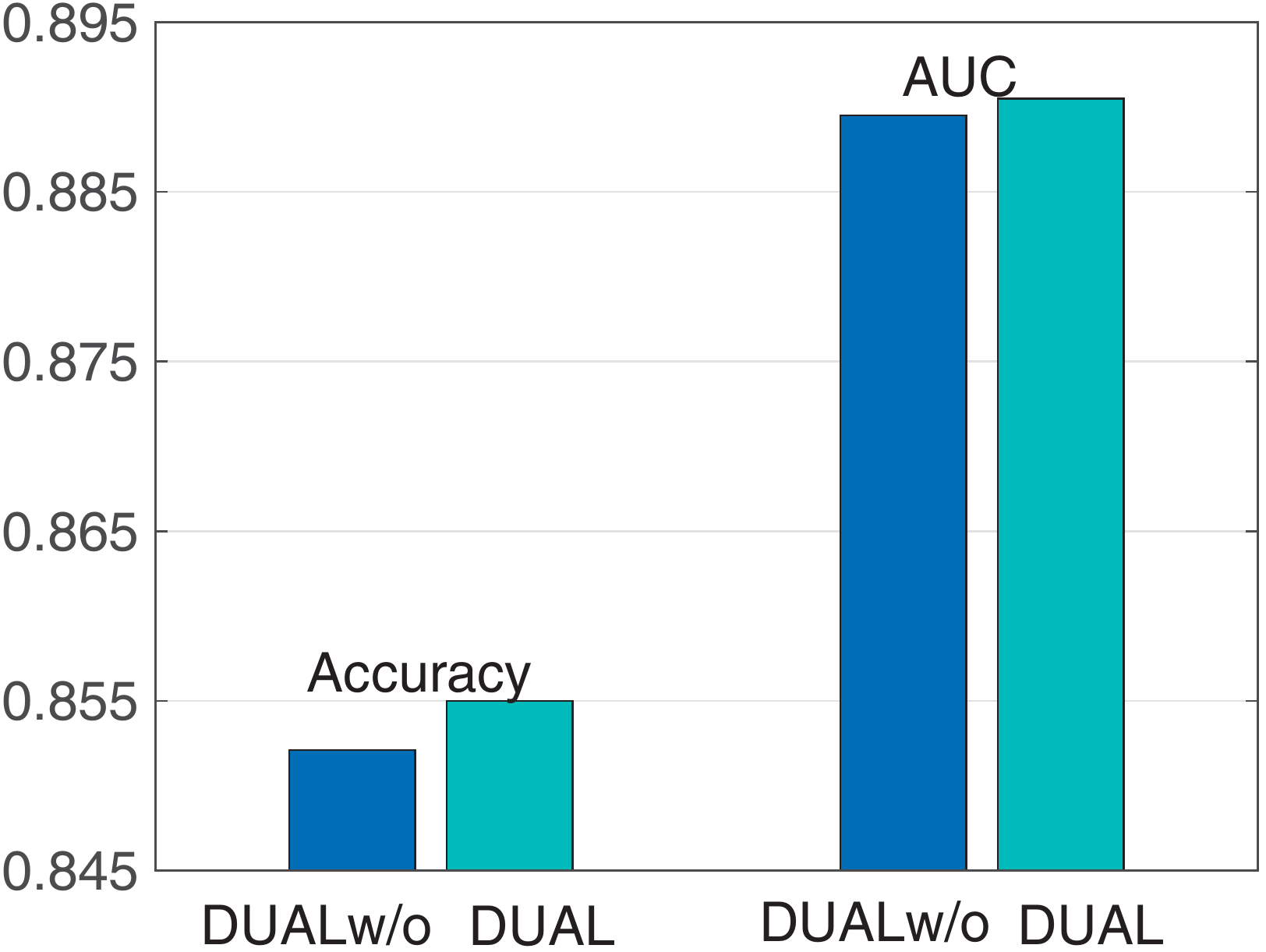}}
\caption{The effectiveness verification of the components in the selection block.}
\label{comp}
\end{figure}
{\textbf{General Performance:}} Figure \ref{accuracy} and \ref{auc} show the scores in terms of different numbers of queries. Our method outperforms other algorithms under most of cases, especially on the larger datasets. This illustrates that by preserving input patterns and cluster structures, DUAL can select representative samples. In addition, we observe that MAED and DCS have good results on some datasets. This may be because MAED is a nonlinear method that can handle complex data, while DCS selects a subset with largest leverage scores, resulting in a good low-rank matrix surrogate.
\begin{figure}
\centering
\includegraphics[width=0.45\linewidth]{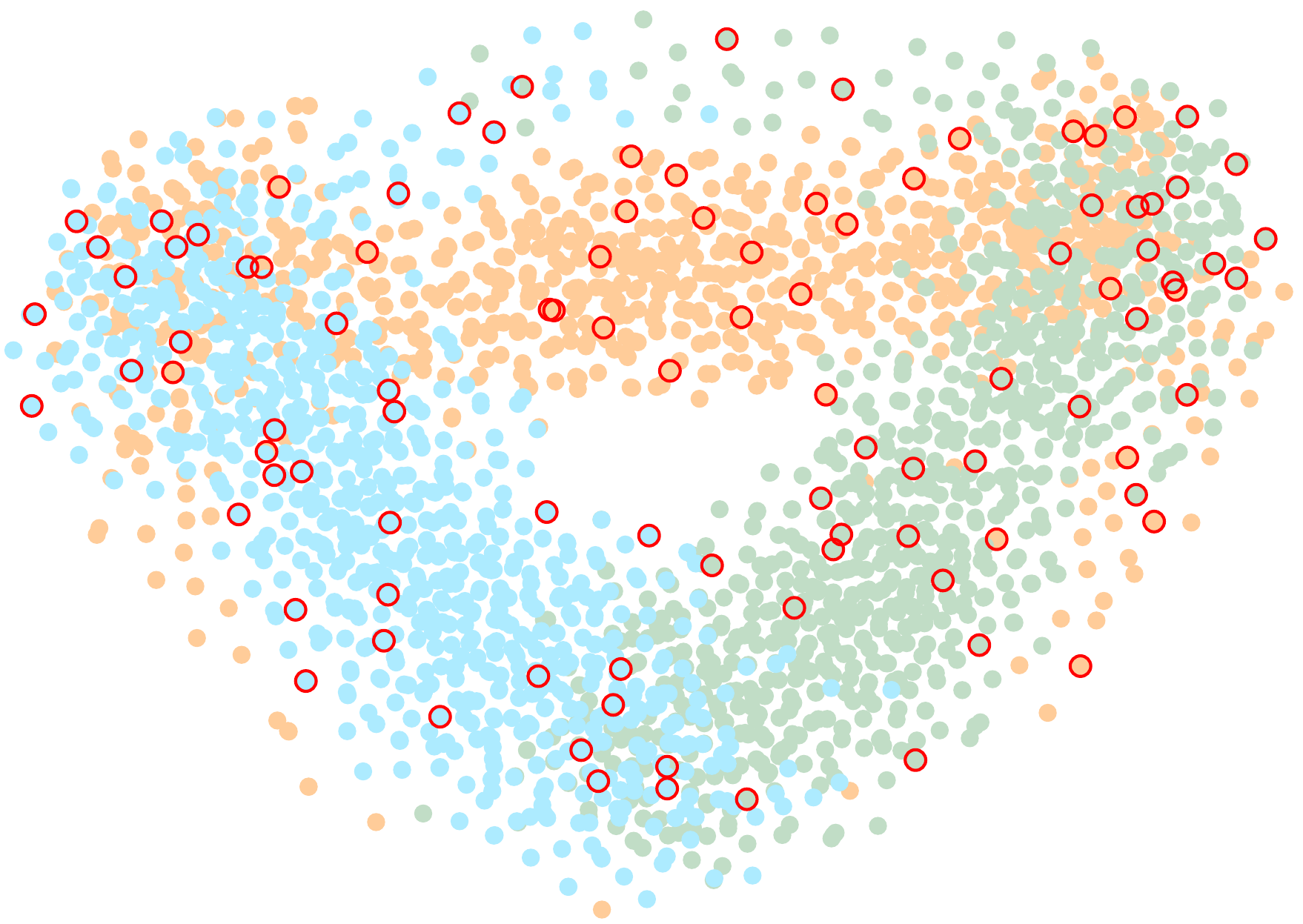}
\caption{The visualization by t-SNE.
The red circles denote the selected samples, and other color solid circles denote different classes. }
\label{visual}
\end{figure}

{\textbf{Ablation Study:}} We study the effectiveness of the components in our selection block on the Urban and Waveform datasets. The experimental setting is as follows: we only consider to preserve the whole input patterns, i.e., setting  $\beta=0$ in (\ref{total loss}). We call it DUALw/o. The number of queries is set to half of the candidates. Figure \ref{comp} shows the results.
 DUAL achieves better results than DUALw/o, which indicates that preserving cluster structures is helpful for unsupervised active learning. On the Waveform dataset, the improvement of DUAL over DUALw/o is a little bit light. This is because the numbers of samples among different classes are balanced in this dataset, and thus selecting samples by only modeling input patterns may preserve the cluster structure well.

{\textbf{Visualization:}}
In this subsection, we apply our method DUAL on the Waveform dataset to give an intuitive result. We use the t-SNE \cite{maaten2008visualizing} to visualize the samples selected by our method, as shown in Figure \ref{visual}.
The samples selected by DUAL can better represent the dataset. This is because DUAL can better capture the
nonlinear structure of data by performing active learning in the latent space.

\begin{figure}
\centering
\includegraphics[width=0.6\linewidth]{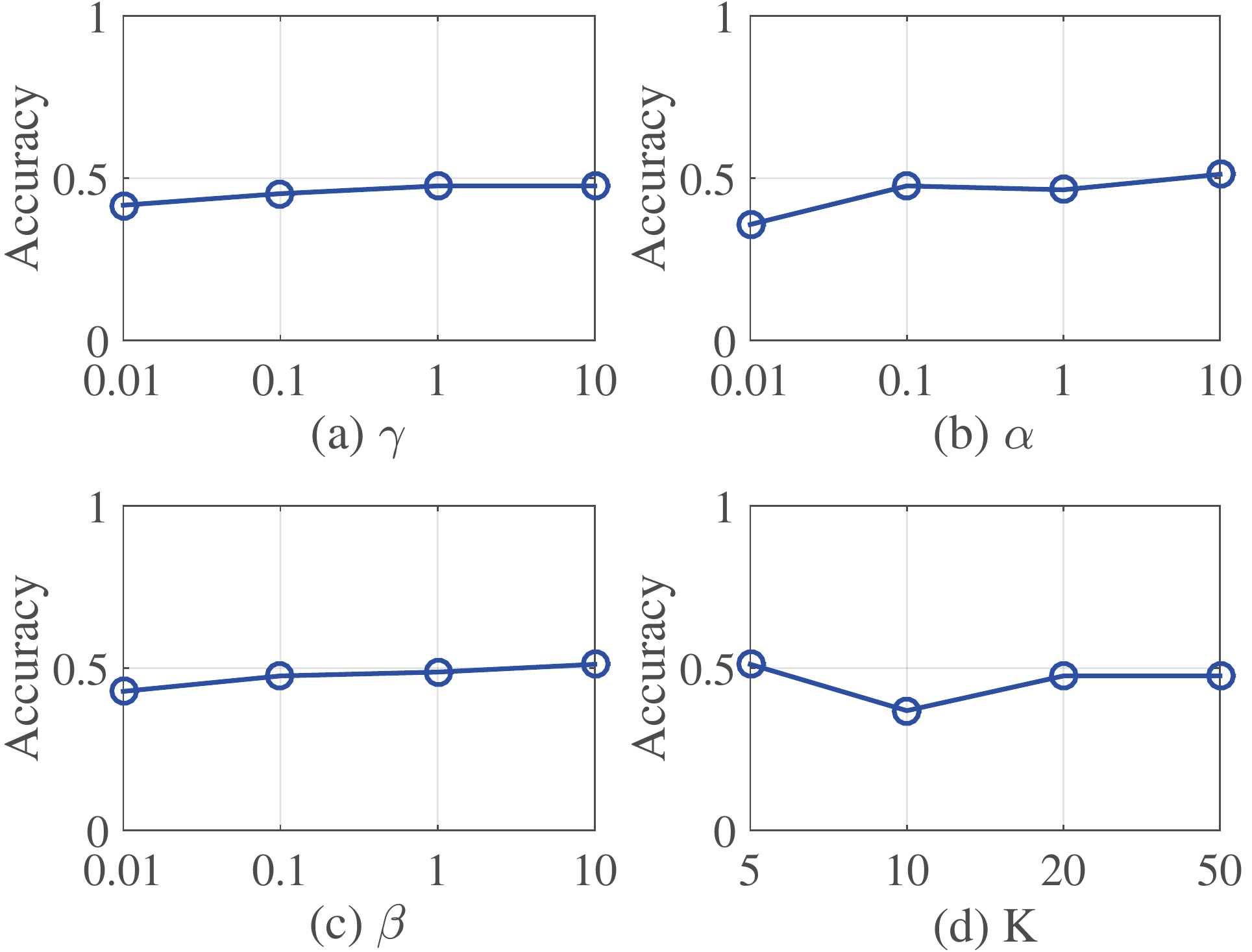}
\caption{Parameter Study on the Urban dataset.}
\label{para}
\end{figure}

{\textbf{Parameter Study:}}
We study the sensitivity of our algorithm about the tradeoff parameters $\alpha$, $\beta$, $\gamma (=\eta)$, and the number of clusters $K$ on the Urban dataset.
We fix the number of queries to half of the candidates, and report their accuracies.
The results are shown in Figure \ref{para}.
Our method is not sensitive to the parameters with a relatively wide range.

\section{Conclusion}
In this paper, we proposed a deep learning based framework for unsupervised active learning, called DUAL. DUAL can model the nonlinear structure of data, and select representative samples to preserve the input pattern and the cluster structure of data. Extensive experimental results on six datasets demonstrate the effectiveness of our method.

\section*{Acknowledgements}

 This work was supported by the NSFC Grant No. 61806044, 61932004, N181605012, and 61732003.

\bibliographystyle{named}
\bibliography{ijcai20}

\end{document}